# INTER-RATER AGREEMENT STUDY ON READABILITY ASSESSMENT IN BENGALI

Shanta Phani[1], Shibamouli Lahiri[2] and Arindam Biswas[3]

[1]Information Technology, IIEST, Shibpur,Howrah 711103, West Bengal, India
[2]Computer Science and Engineering, University of North Texas, Denton, TX 76207, USA
[3]Information Technology, IIEST, Shibpur,Howrah 711103, West Bengal, India

## ABSTRACT

*An inter-rater agreement study is performed for readability assessment in Bengali. A 1-7 rating scale was used to indicate different levels of readability. We obtained moderate to fair agreement among seven independent annotators on 30 text passages written by four eminent Bengali authors. As a by product of our study, we obtained a readability-annotated ground truth dataset in Bengali.*

## 1. INTRODUCTION

Readability refers to the ease with which a given piece of natural language text can be read and understood. Intuitively, readability emerges from an interaction between the reader and the text, and depends on the prior knowledge of the reader, his/her reading skills, interest, and motivation [23]. Although it may seem that automatic assessment of readability would be a very complicated process, as it turns out, fairly effective readability scoring can be achieved by means of several lowlevel features.

Readability has many important applications, such as assessing the quality of student essays (one of the original applications of readability scoring), designing educational materials for schoolchildren and second-language learners, moderating newspaper content to convey information more clearly and effectively, and standardizing the language-learning experience of different age groups. Readability ("reading ease") and its converse – reading difficulty – are associated with different grade levels in school. It is generally observed that students from higher grade levels can write and comprehend texts with greater reading difficulty than students from lower grade levels. A lot of studies in readability therefore focused on correlating readability scores with grade levels, or even predicting grade levels from readability-oriented features. Existing methods of readability assessment look into a handful of low-level signals such as average sentence length (ASL), average word length in syllables (AWL), percentage of difficult words, and number of polysyllabic words. Early studies used word frequency lists to identify difficult words. Recently, readability evaluation has been tackled as a supervised machine learning problem [12], [17], [22].

There have been many different studies on readability assessment in English (cf. Section 2). Bengali has received much less attention owing to inadequate resources and a lack of robust natural language processing tools. It is only very recently that some groups of researchers looked into readability assessment in Bengali. They observed that English readability formulas did not work well on Bengali texts [11], [21]. This observation is not surprising, because Bengali is very different than English. Bengali is a highly inflected language, follows subject-object-verb ordering in sentences, and has a rich morphology. Further, Bengali shows word compounding and diglossia, i.e. formal and informal language variants (*sadhu bhasha* and *cholit bhasha*). All these factors complicate readability scoring in Bengali. Since the concept of readability is highly subjective and reader-dependent, it is necessary to find out how much two native Bengali speakers agree on the readability level of a piece of text. Generalizing from there, we performed an inter-rater agreement study on readability assessment in Bengali. This study not only enables us to see how much human annotators agree on readability assessment, but also shows how difficult it is for humans to assign consistent readability scores. Since Bengali is very different than English, we want to see if (and how) readability is affected by the peculiarities of the language. As a by-product of this study, we obtained a human-annotated gold standard dataset for readability evaluation in Bengali. The rest of this paper is organized as follows. We briefly discuss related studies in Section 2, followed by a discussion of our dataset and annotation scheme in Section 3. Experimental results are described in Section 4, along with their explanation and observations. Section 5 concludes the paper with contributions, limitations, and further research directions.

Table 2. Mean Readability Rating and Standard Deviation of 30 Text Passages

| Mean Rating | Standard Deviation |
|:---:|:---:|
| 2.0 | 0.52 |
| 2 | 0.75 |
| 2 | 1.26 |
| 2 | 1.17 |
| 3 | 0.98 |
| 4 | 0.75 |
| 2 | 1.10 |
| 4 | 1.47 |
| 4 | 1.26 |
| 4 | 1.05 |
| 5 | 1.05 |
| 3 | 0.98 |
| 3 | 1.33 |
| 4 | 1.52 |
| 3 | 0.75 |
| 4 | 1.60 |
| 4 | 1.50 |
| 4 | 1.72 |
| 3 | 1.03 |
| 5 | 1.76 |
| 4 | 1.17 |
| 3 | 1.03 |
| 5 | 1.75 |
| 4 | 1.67 |
| 3 | 1.55 |
| 3 | 0.84 |
| 6 | 1.50 |

| 5 | 1.50 |
|---|------|
| 5 | 1.17 |
| 4 | 1.03 |

## 2. RELATED WORK

Readability scoring in English has a long and rich history, starting with the work of L. A. Sherman in the late nineteenth century [20]. Among the early readability formulas were Flesch Reading Ease [7], Dale-Chall Formula [5], Automated Readability Index [19], Gunning Fog Index [9], SMOG score [16], and Coleman-Liau Index [2]. These early indices were based on simple features like average number of characters, words, syllables and sentences, number of difficult and polysyllabic words, etc. Albeit simple, these readability indices were surprisingly good predictors of a reader's grade level. Two different lines of work focused on children and adult readability formulas. Recently Lahiri et al. showed moderate correlation between readability indices and formality score ([10]) in four different domains [14].

Sinha et al. classified English readability formulas into three broad categories – traditional methods, cognitively motivated methods, and machine learning methods [21]. Traditional methods assess readability using surface features and shallow linguistic features such as the ones mentioned in the preceding paragraph. Cognitively motivated methods take into account the cohesion and coherence of text, its latent topic structure, Kintsch's propositions, etc [1], [8], [13]. Finally, machine learning methods utilize sophisticated structures such as language models [3], [4], [18], query logs [15], and several other features to predict the readability of open- domain text data.

There are very few studies on readability assessment in Bengali texts. We found only three lines of work that specifically looked into Bengali readability [6], [11], [21]. Das and Roychoudhury worked with a miniature model of two parameters in their pioneering study [6]. They found that the two-parameter model was a better predictor of readability than the one-parameter model. Note, however, that Das and Roychoudhury's corpus was small (only seven documents), thereby calling into question the validity of their results. Sinha et al. alleviated these problems by considering six parameters instead of just two [21]. They further showed that English readability indices were inadequate for Bengali, and built their own readability model on 16 texts. Around the same time, Islam et al. independently reached the same conclusion [11]. They designed a Bengali readability classifier on lexical and information-theoretic features, resulting in an F-score 50% higher than that from traditional scoring approaches.

While all the above studies are very important and insightful, none of them explicitly performed an inter-rater agreement study. For reasons mentioned in Section 1, an inter-rater agreement study is very important when we talk about readability assessment. Further, none of these studies made available their readability-annotated gold standard datasets, thereby stymieing further research. We attempt to bridge these gaps in our work.

## 3. METHODOLOGY

We collected a corpus of 30 Bengali text passages. The passages were randomly selected from the writings of four eminent Bengali authors – Rabindranath Tagore (1861-1941), Sarat Chandra Chattopadhyay (1876-1938), Bankim Chandra Chattopadhyay (1838-1894), and Bibhutibhushan Bandyopadhyay (1894-1950). We ensured that samples from both sadhu bhasha as well as cholit bhasha were incorporated in our corpus. We also ensured that we had both adult text as well as children's text in the mix. The number of passages from different authors is shown in Table 1.

Table 1 also shows the number of passages in sadhu bhasha and cholit bhasha. Note that there are almost twice as many passages in sadhu bhasha as in cholit bhasha.

Table 1: Number of Bengali text passages from different Authors, and from two different bengali language forms: Sadhu bhasha and Cholit bhasha

| Author | Sadhu Bhasha | Chalit Bhasha | Total |
|---|---|---|---|
| Rabindranath Tagore | 8 | 4 | 12 |
| Sarat Chandra Chattopadhyay | 6 | 3 | 9 |
| Bankim Chandra Chattopadhyay | 6 | 0 | 6 |
| Bibhutibhusan Bandopadhyay | 1 | 2 | 3 |
| Total | 21 | 9 | 30 |

We assigned the 30 text passages to seven independent annotators. The annotators were 30 to 35 years of age; they were from a similar educational background and socio-economic milieu; there were four female and three male annotators; and they all were native speakers of Bengali. Annotators were asked to assign a readability rating to each of the 30 passages.
The rating scale was as follows:
1) Very easy to read
2) Easy to read
3) Somewhat easy to read
4) In-between
5) Somewhat difficult to read
6) Difficult to read
7) Very difficult to read

This rating scale reflects the fact that readability is not a binary/ternary variable; it is an ordinal variable. We further collected the data on whether the annotators were avid readers of Bengali or not. Each annotator rated every passage. Note that readability annotation in Bengali is challenging because passages written in sadhu bhasha tend to be harder to read than those written in cholit bhasha. Since our dataset contains both sadhu bhasha and cholit bhasha, maintaining consistency in readability rating becomes a big issue.

## 4. RESULTS

Table 2 gives the mean readability rating of the 30 text passages, along with their standard deviations. These ratings are averages over seven independent annotations. Note from Table 2 that none of the mean ratings is 1 or 7. In other words, mean ratings never reach the extreme readability values. This phenomenon is known as the central tendency bias. Note also that the standard deviations are not very high, which should be intuitive because the rating scale varies between 1 and 7.Agreement among the annotators was measured by Cohen's kappa ($\kappa$) and Spearman's rank correlation coefficient ($\rho$). Table 3 shows the pairwise $\kappa$ values among different annotators, and Table 4 gives the pairwise $\rho$ values. Both tables are symmetric around the main diagonal. Note from Table 3 that 11 out of 21 $\kappa$ values fall within the range [0:5; 0:8]. Table 4 shows that 13 out of 21 $\rho$ values are within the range [0:5; 0:8], and one $\rho$ value is greater than 0.8. This indicates moderate to fair agreement among different annotators. This observation in turn indicates that human annotators agree pretty well on Bengali readability scoring.

Table 3. Cohen's Kappa Between Different Annotators

|  | Annotator1 | Annotator2 | Annotator3 | Annotator4 | Annotator5 | Annotator6 | Annotator7 |
|---|---|---|---|---|---|---|---|
| **Annotator1** | 1.00 | 0.51 | 0.26 | 0.60 | 0.12 | 0.51 | 0.50 |
| **Annotator2** | 0.51 | 1.00 | 0.45 | 0.64 | 0.12 | 0.56 | 0.53 |
| **Annotator3** | 0.26 | 0.45 | 1.00 | 0.43 | -0.03 | 0.49 | 0.54 |
| **Annotator4** | 0.60 | 0.64 | 0.43 | 1.00 | 0.09 | 0.58 | 0.74 |
| **Annotator5** | 0.12 | 0.12 | -0.03 | 0.09 | 1.00 | 0.13 | 0.02 |
| **Annotator6** | 0.51 | 0.56 | 0.49 | 0.58 | 0.13 | 1.00 | 0.57 |
| **Annotator7** | 0.50 | 0.53 | 0.54 | 0.74 | 0.02 | 0.57 | 1.00 |

Table 4. Spearman Rank Correlation between Different Annotators

|  | Annotator1 | Annotator2 | Annotator3 | Annotator4 | Annotator5 | Annotator6 | Annotator7 |
|---|---|---|---|---|---|---|---|
| **Annotator1** | 1.00 | 0.62 | 0.34 | 0.70 | 0.16 | 0.60 | 0.60 |
| **Annotator2** | 0.62 | 1.00 | 0.52 | 0.75 | 0.14 | 0.65 | 0.63 |
| **Annotator3** | 0.34 | 0.52 | 1.00 | 0.53 | -0.04 | 0.59 | 0.64 |
| **Annotator4** | 0.70 | 0.75 | 0.53 | 1.00 | 0.12 | 0.68 | 0.83 |
| **Annotator5** | 0.16 | 0.14 | -0.04 | 0.12 | 1.00 | 0.15 | 0.02 |
| **Annotator6** | 0.60 | 0.65 | 0.59 | 0.68 | 0.15 | 1.00 | 0.66 |
| **Annotator7** | 0.60 | 0.63 | 0.64 | 0.83 | 0.02 | 0.66 | 1.00 |

# 5. CONCLUSION

We performed an inter-rater agreement study for readability assessment in Bengali. This is the first time such an agreement study has been performed. We obtained moderate to fair agreement among seven independent annotators on 30 text passages written by four eminent Bengali authors. As a byproduct of this study, we obtained a gold standard human annotated readability dataset for Bengali. We plan to release this dataset for future research. We are working on readability modeling in Bengali, and this dataset will be very helpful. An important limitation of our study is the small corpus size. We only have 30 annotated passages at our disposal, whereas Islam et al. [11] had around 300. But Islam et al.'s dataset is not annotated in as fine-grained a fashion as ours. Note also that our dataset is larger than both Sinha et al.'s 16document dataset [21], and Das and Roychoudhury's seven document dataset [6]. We plan to increase the size of our dataset in future.

**Authors**
Shanta Phani: She is a PhD student in IIEST, Kolkata. She has done her M.Tech in Computer Science and Engineering, from West Bengal University of Technology, in 2009. Her research interests are in image processing, Natural Language Processing, especially Authorship Attribution, style identification, etc for Bengali language.

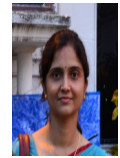

Shibamouli Lahiri: He is currently a PhD student at University of North Texas. He completed his Master of Engineering from Pennsylvania State University in 2012. His primary research interests are in Natural Language Processing, Data Mining and Machine Learning. He has worked in many different problems in NLP, including but not limited to keyword extraction, multi-document summarization, formality analysis, authorship attribution, and native language identification. His research involves the application of machine learning and complex network analysis techniques to different problems in Natural Language Processing.

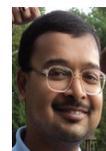

Dr. Arindam Biswas: He has done is doctors from ISI Kolkata. He has over 50 published papers. He works in the area of Digital Geometry, Medical Image Analysis, and Natural Language Processing.

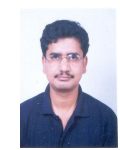